\documentclass{article}

\PassOptionsToPackage{numbers, compress}{natbib}

\usepackage[preprint]{neurips_2024}

\usepackage{amsmath}
\usepackage[utf8]{inputenc} 
\usepackage[T1]{fontenc}    
\usepackage{hyperref}       
\usepackage{cleveref}		
\usepackage{url}            
\usepackage{booktabs}       
\usepackage{amsfonts}       
\usepackage{nicefrac}       
\usepackage{microtype}      
\usepackage{xcolor}         
\usepackage{graphicx}
\usepackage{array}
\usepackage{booktabs} 
\usepackage{tabularx}
\usepackage{multirow}
\usepackage{ulem}
\title{OpFlowTalker: Realistic and Natural Talking Face Generation via Optical Flow Guidance}

\author{%
  Shuheng Ge, Haoyu Xing, Li Zhang, Xiangqian Wu\thanks{Use footnote for providing further information about author (webpage, alternative address)---\emph{not} for acknowledging
    funding agencies.} \\
  Harbin Institute of Technology \\
  China\\
  \texttt{\{suheng,xqwu\}@hit.edu.cn} \\
}

\begin{document}

\maketitle

\begin{abstract}
\label{Abstract}
  Creating realistic, natural, and lip-readable talking face videos remains a formidable challenge. Previous research primarily concentrated on generating and aligning single-frame images while overlooking the smoothness of frame-to-frame transitions and temporal dependencies. This often compromised visual quality and effects in practical settings, particularly when handling complex facial data and audio content, which frequently led to semantically incongruent visual illusions. Specifically, synthesized videos commonly featured disorganized lip movements, making them difficult to understand and recognize. To overcome these limitations, this paper introduces the application of optical flow to guide facial image generation, enhancing inter-frame continuity and semantic consistency. We propose "OpFlowTalker", a novel approach that predicts optical flow changes from audio inputs rather than direct image predictions. This method smooths image transitions and aligns changes with semantic content. Moreover, it employs a sequence fusion technique to replace the independent generation of single frames, thus preserving contextual information and maintaining temporal coherence. We also developed an optical flow synchronization module that regulates both full-face and lip movements, optimizing visual synthesis by balancing regional dynamics. Furthermore, we introduce a Visual Text Consistency Score (VTCS) that accurately measures lip-readability in synthesized videos. Extensive empirical evidence validates the effectiveness of our approach.
\end{abstract}
\section{Introduction}
\label{Introduction}
Due to its immense potential in various application contexts such as virtual reality\cite{peng2023selftalk}, film production\cite{kim2018deep}, and online education\cite{pataranutaporn2021ai}, talking face generation has attracted increasing attention. 
Recent research has primarily focused on the restoration of facial features\cite{tan2024flowvqtalker,tan2024style2talker}, head movement\cite{zhang2023sadtalker}, and the synthesis of expressions\cite{otroshi2024face,yang2024pgdiff}. However, studies on lip synchronization remain in the early stages\cite{prajwal2020lip}.
Lip synchronization is a critical indicator for evaluating the quality of talking face generation.
Although existing research has proposed various enhancements for synchronizing single-frame images with audio encoding\cite{chen2019hierarchical,cheng2022videoretalking,kr2019towards,zhou2021pose,pataranutaporn2021ai}, it has neglected the smoothness of consecutive frame transitions and the readability of lip movement changes. These aspects are crucial components in creating a realistic and natural virtual avatar.

\begin{figure}[htbp]
	\centering
	\includegraphics[width=\linewidth]{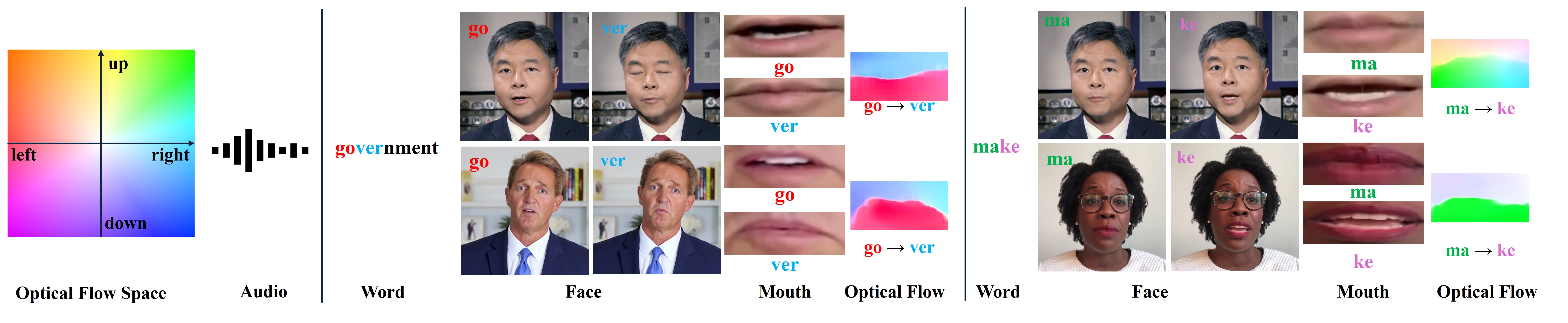}
	\caption{The example shows faces, lip shapes, and lip optical flow for different scenarios: different individuals producing the same vocalization and the same individual making different vocalizations. On the left, there is a visualization of the optical flow space, where different colors represent different directions of optical flow, and the depth of color indicates the intensity of the flow. }
	\label{fig:flow_example}
\end{figure}
To address this issue, we conducted an analysis of talking face videos from the perspectives of kinematics and linguistics \cite{gibson1950perception}. This investigation led to the identification of three important pieces of information: 1) The visual changes in sequential facial images should comply with a smoothly regulated optical flow field. Specifically, under constant brightness conditions, facial movements between consecutive frames are minor\cite{ilg2017flownet}. Changes over time should not cause drastic alterations in facial information, particularly in the movements of the lips. 2) According to linguistic and biological principles\cite{roach2001phonetics,son2017lip}, the variations in lip movements during the articulation of identical or similar syllables by different individuals should be consistent or closely similar, thus following the same optical flow field. This optical flow field remains unaffected by environmental changes \cite{ma2022visual}. As shown in \cref{fig:flow_example}, the optical flow of the lips for different individuals is nearly identical under the sequence of the same audio syllables, which is fundamental to achieving a realistic visual representationoo. 3) For the same individual speaking different content, the differences in the images are not as pronounced as the differences in the optical flow. This observation indicates that Optical flow effectively highlights the subtle changes in lip motions that correspond to different phonetic articulations, providing a more detailed and nuanced understanding of how speech variations influence facial dynamics. 

For synchronizing audio and visual elements, previous research typically achieved alignment by matching single-frame imagery with discrete audio encodings, and independently deducing facial information and lip features for each segment from the audio sequence. \cite{prajwal2020lip,zhang2023sadtalker} involved training an Expert Discriminator to align lips by calculating the average distance between generated images and real images. Meanwhile, \cite{peng2023synctalk} entailed pre-training an audio-visual encoder on a customized character dataset to replace traditional audio features, thereby enforcing the alignment between single-frame images and audio features. While these approaches can alleviate the issue of audio-lip mismatch to some extent, they still present several limitations: 1) The generation of single-frame images occurs independently, neglecting the coherence and dependency of temporal changes. 2) By emphasizing the average distance of image sequences, they force frequent and rapid changes in lip movement within a very short period. 3) They overlook the consistency of lip movements for identical or similar syllables, and the customization involved in pre-training encoders reduces their general applicability. Overall, these methods do not sufficiently consider the three critical pieces of information mentioned earlier.

In this paper, we propose a novel method called OpflowTalker to generate realistic and natural talking face videos with fluent and readable lip movements. OpFlowTalker offers the following advantages: (1) It robustly generates diverse talking face videos for different characters and datasets of varying quality. (2) The synthesized videos are smoother and more natural, especially in terms of lip movements. (3) The synthesized talking face videos demonstrate exceptionally high performance in lip-reading recognition. OpFlowTalker includes two specialized components: the Facial Sequential Generation via Optical Flow (FSG) and the Optical Flow Synchronization Module (OFSM). Within the FSG, we embed audio features into a gaussian distribution \cite{zhang2018stackgan++} and replace the original audio features \cite{he2016deep,prajwal2020lip} with random sampling from the distribution as the input for predictive generation, enhancing the diversity and robustness of generated face. The optical flow predictor randomly samples from a noise distribution and learns to predict facial optical flow based on prior facial features and the currently sampled audio features by using MLPs. The generator reconstructs the current facial image through the predicted facial optical flow and prior facial features, achieving serialized synthesis of facial information. At the same time, the original reference image is also appended to the facial features to maintain consistency of ID information.

On the other hand, OFSM is designed to supervise the alignment of audio-lip synchronization and smooth facial changes. We decompose facial feature information into invariant identity information and variable sequence information, with the invariant identity information maintained by calculating the facial reconstruction loss. And the variable sequence information is constrained by calculating the average optical flow loss \cite{ilg2017flownet,jiang2021learning,teed2020raft} across the same consecutive frames. Specifically, to avoid the detailed lip motion being overly smoothed by the general facial information, and to balance fluidity and consistency, we divide the optical flow loss into facial optical flow and lip optical flow. These are introduced at different stages of training to ensure high fidelity and expressiveness of the synthesized lip information. The lip region is captured using facial keypoints \cite{bulat2022pre}. Additionally, we propose a novel evaluation metric based on the lip-reading model \cite{ma2022visual}, the Visual Text Consistency Score (VTCS), to assess the restoration degree and readability of synthesized visual facial videos in lip-syncing. In summary, our main \textbf{contributions} are as follows:
\begin{itemize}
	\item We propose a novel talking face generation method, OpFlowTalker, which generates fluent, natural, and highly lip-reading intelligibility videos by predicting inter-frame optical flow from audio, replacing the direct prediction of images. 
	\item We have systematically applied optical flow for the first time in the task of talking face generation, and we propose sequence fusion prediction and optical flow consistency loss to enhance the inter-frame continuity and semantic consistency of synthesized videos.
	\item we propose the Visual Text Consistency Score (VTCS), a novel evaluation metric to measure the lip-reading intelligibility of synthesized videos. The effectiveness of VTCS has been validated through the results of multiple methods.
	\item Extensive qualitative and quantitative experiments on two challenging datasets  that our method achieves state-of the-art performance and exhibits excellent generalizability.
\end{itemize}

\section{Related Works}
\label{Related Works}

\subsection{Audio-Driven Talking Face Generation}
Audio-driven talking face generation \cite{bregler2023video,liu2022semantic} can be broadly categorized into customized-synthesis methods and general-synthesis methods.  Customized-synthesis methods \cite{du2023dae, guo2021ad, peng2023synctalk, suwajanakorn2017synthesizing} leverage extensive data of specific individuals to train end-to-end models that synthesize lip-synced videos from audio input. These methods typically require a large amount of specific person video data to achieve good results and lack good generalization on other character datasets. General-synthesis methods achieve generalizability for generating different characters by learning intermediate representations \cite{chen2020talking,das2020speech,wang2021audio2head,wang2022one,zakharov2019few,zhang2023sadtalker,zhong2023identity,zhou2020makelttalk} or reconstruction \cite{chen2018lip,kr2019towards,prajwal2020lip,shen2022learning,shen2023difftalk,song2018talking,thies2020neural,wang2021audio2head,zhou2019talking}. ATVGnet \cite{chen2019hierarchical} uses the facial landmark as the intermediate representation to generate the video frames. Similarly, MakeItTalk \cite{zhou2020makelttalk} uses 2D facial key points as an intermediate representation, but it attempts to decouple speech content and identity information from the input audio signal. To address the issue of unnatural head movements and facial expressions due to the lack of 3D information, PC-AVS \cite{zhou2021pose} disentangles the head pose and expression using 3DMM parameters \cite{blanz2023morphable}. Some methods \cite{zhang2023sadtalker,tan2024flowvqtalker} have further optimized along this line of thought. These methods usually require an additional information as the input to control motion changes. The extraction of intermediate representations is time-consuming and laborious. And they often provide limited facial dynamics details. Due to these reasons, these methods have limited practical applicability. Reconstruction-based methods offer better generative results. Wav2lip \cite{pataranutaporn2021ai} employs a pre-trained lip-sync expert discriminator to enhance the accuracy of lip movements, which has been used by the majority of subsequent research \cite{cheng2022videoretalking,zhang2023sadtalker}. However, because the discriminator overly emphasizes the alignment of single-frame lip movements within a window, when dealing with non-uniform audio, the synthesis of continuous lip images often results in hallucinations, producing lip shapes that do not match the semantics. This problem has not been adequately addressed in subsequent related work \cite{cheng2022videoretalking,wang2023seeing,zhang2023sadtalker}.

\subsection{Optical flow estimation}
The concept of optical flow was first introduced by Gibson in 1950 \cite{gibson1950perception}. It describes the instantaneous velocity of pixel movement caused by moving objects in space on the imaging plane. The purpose of optical flow estimation is to accurately capture the motion and changes of every pixel between consecutive frames. Optical flow estimation can be divided into two types: sparse optical flow and dense optical flow. Sparse optical flow typically involves tracking only a subset of points in an image, while dense optical flow involves estimating the motion of every pixel in the image. Optical flow estimation is categorized into traditional image-based optical flow computation \cite{horn1981determining,lucas1981iterative} and deep network-based optical flow prediction \cite{dosovitskiy2015flownet,ilg2017flownet,sun2018pwc,teed2020raft,jiang2021learning}. FlowNet \cite{dosovitskiy2015flownet} formulates the optical flow estimation problem as a supervised learning task and constructs a convolutional neural network capable of autonomously learning optical flow. This approach not only improves the accuracy of optical flow estimation but also speeds up the computation, making it more practical for real-world applications. RAFT \cite{teed2020raft} achieves outstanding generalization performance using Recurrent All-Pairs Field Transformers and runs at a high frame rate. GMA tackles the problem of estimating optical flow for occluded points by modeling image self-similarity and leveraging global motion aggregation modules and transformer techniques.
Optical flow inherently reflects the temporal continuity and motion smoothness of the content in videos. To make the generated videos more natural and fluid, we have introduced optical flow prediction and the calculation of optical flow estimation loss into talking face video generation for the first time.

\begin{figure}[htbp]
	\centering
	\includegraphics[width=\linewidth]{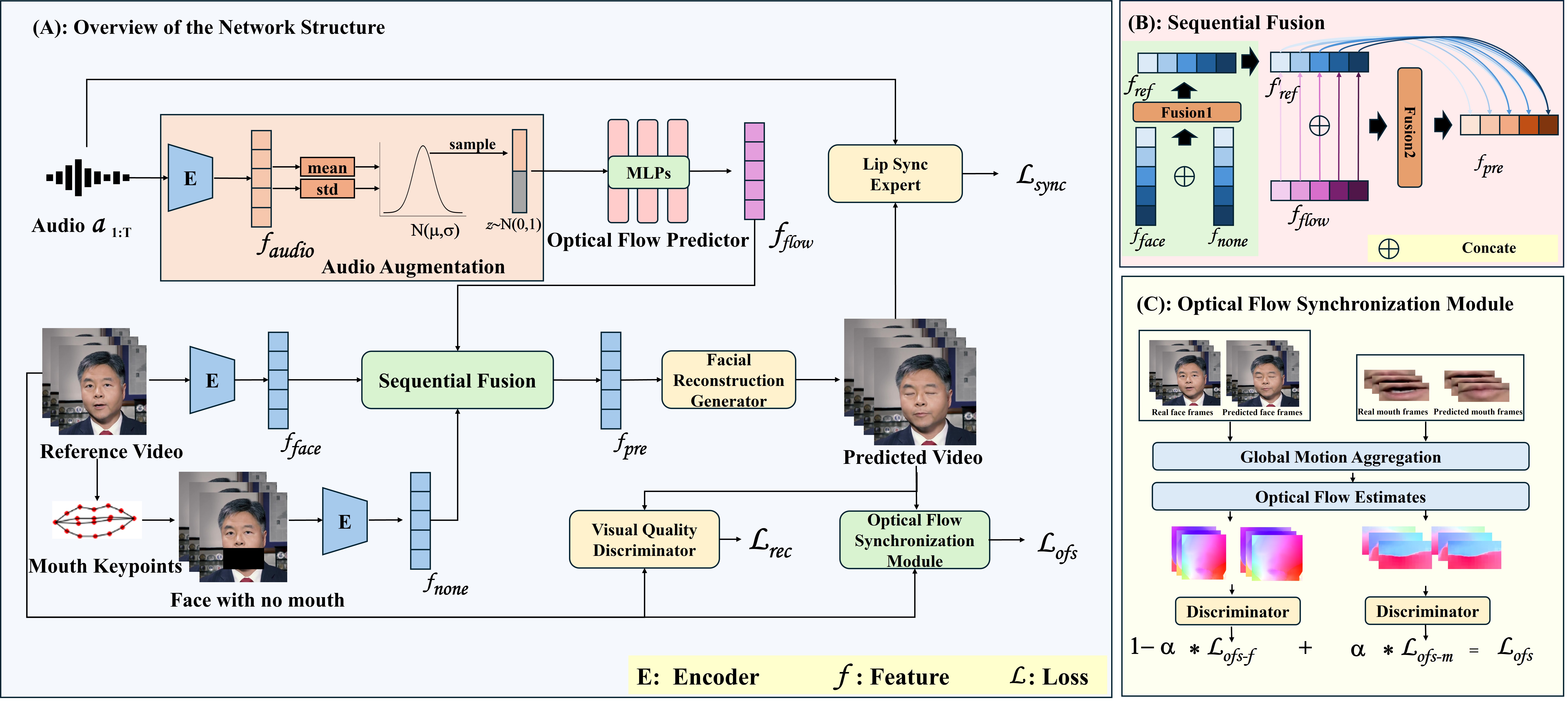}
	\caption{\text{llustration of our proposed OpFlowTalker}. (a) \textbf{OpFlowTalker framework}, which generates talking faces based on consistent audio input $a$. OpFlowTalker integrates four effective components to enhance the generated videos quality: Audio Augementation, Optical Flow Predictor, Sequential Fusion, Optical Flow Synchronization Module. (b) \textbf{Sequential Fusion}, which predictes each frame relying on all preceding reference information.Fusion1 consists of 2-linear-layers and Fusion2 is a 6-layer Transformer.  (c)\textbf{Optical Flow Synchronization Module}, which aggregates global motion information and calculates the synchronization loss of facial and lip optical flow separately. $\alpha$ is generally set to 0.1. }
	\label{fig:network}
\end{figure}
\section{Methodology}
\label{Methodology}
FlowTalker adheres to the fundamental structure of the Wav2Lip model \cite{prajwal2020lip} , incorporating newly designed prediction and supervision modules. In this section, we will first introduce the Facial Sequential Generation via Optical Flow in Section 3.1. Subsequently, we will provide a detailed discussion of the Optical Flow Synchronization Module in Section 3.2.
\subsection{Facial Sequential Generation via Optical Flow}
\paragraph{Audio augmentation}
Currently, the processing of audio employs the use of the Mel Spectrogram for signal extraction, followed by the utilization of an encoder for audio embeddings which are then directly used as conditional vectors for subsequent tasks. Previous research \cite{reed2016generating,zhang2018stackgan++} indicates that the dimensionality of the encoding space is higher compared to embeddings of semantic information. Utilizing encoded information directly as a condition may lead to a sparse vector distribution in the latent space, which is not desirable for learning the generator. Moreover, the use of fixed audio features also limits the diversity and robustness of the generated output. To mitigate this problem, we introduce a Audio Augmentation Module to adjust the audio embeddings. The audio signal $a$ processed by the Mel Spectrogram is sent to the audio encoder \cite{amodei2016deep,prajwal2020lip} to obtain the audio feature $f_{audio}$. For each audio segment feature, we construct an independent Gaussian distribution $N(\mu ,\sigma)$, where the mean $\mu$ and diagonal covariance matrix $\sigma$ are functions of the audio feature $f_{audio}$. We randomly sample a latent variable from the audio distribution and concatenate it with a noise to form a new set of audio features $f^{\prime}_{audio}$. 

\paragraph{Optical Flow Predictor} 
Instead of directly generating facial images, we predict a set of optical flows from audio features, which sequentially guide the face motion of the reference image. For a video of duration $T$, we define the audio sequence as $A = \{a_1, a_2, \dots, a_t\}$ and the frame sequence as $I= \{i_0, i_1, i_2, \dots, i_t\}$. The optical flow variation between two consecutive frames, $i_m$ and $i_n$, is defined as $o_{m,n}$. It contains comprehensive visual information about facial movements and serves as the foundation for the smooth transitions within facial imagery. For a talking face video:
\begin{equation}
	i_0 \xrightarrow{o_{0,1}} i_1 \xrightarrow{o_{1,2}} i_2 \dots i_{t-1} \xrightarrow{o_{t-1,t}} {i_t} 
\end{equation}
$O = \{o_1, o_2, \dots, o_t\}$ represents a collection of optical flow information encapsulating changes across all video frames, documenting the motion patterns of pixel points within an image sequence, thereby naturally reflecting the temporal continuity and motion smoothness in the video. This enables the prediction of subsequent frame images based on the information from the previous frame. Compared to static facial features, optical flow, as an intermediate representation, can be abstracted from the facial features of specific individuals, enhancing the model's generalization capabilities across different faces and expressions. This is attributable to the fact that optical flow focuses more on patterns of movement rather than the specific facial details of individuals. Additionally, as representations of temporal and semantic information \cite{son2017lip} , optical flow features exhibit a higher correlation with audio features.  exhibit a higher correlation with audio features. Predicting optical flow rather than direct facial features can prove more robust to noise and variations in input data. Minor disturbances in audio signals are unlikely to directly impact the overall pattern of optical flow, which could enhance the model's reliability in real-world applications. Consequently, it is easier to learn optical flow features from audio. To fully preserve the temporal relationships and semantic information and to minimize the increase in model complexity introduced by intermediate modalities, as shown in \cref{fig:network}.(a), we adopt MLPs (Multilayer Pereptron) to perform affine transformations on audio features to predict optical flow features:
\begin{equation}
	f_{flow} = MLPs (AUG(f_{audio}),z) 
\end{equation}
$AUG$ stands for audio augmentation, and $z$ represents randomly sampled Gaussian noise from a standard normal distribution, $N(0 ,1)$ .

\paragraph{Sequential Fusion} 
For the generation of consecutive frames in talking face videos, it is common to embed the audio encoding independently into the reference image to predict each frame's image separately. Particularly in methods that use video as the reference input, the video frames are concatenated along the batch size dimension before encoding and prediction. This approach leads to a loss of correlation and temporal information between generated images. The interaction between predicted images relies solely on layer normalization and loss constraints, often resulting in hallucinatory phenomena in facial information during continuous changes. As shown in \cref{fig:network}.(b), for video frames within a specified window, we have innovatively designed a Sequential Fusion Module that allows predictions from adjacent frames to fully perceive contextual information. This ensures the consistency of facial features and the smoothness of lip movements, thereby reducing the visual hallucinations caused by independent predictions. We employed a keypoint network \cite{bulat2022pre} to extract facial keypoints and identify the lip region, subsequently masking this specified area. Both the complete sequence of reference images and the sequence of masked images were input into an encoder to generate facial feature sequences $f_{face}$ and $f_{none}$. These sequences were concatenated and then fed into Feature Fusion Unit $\phi_1$ to obtain the reference features $f_{ref}$. 
The merged visual features are concatenated with the predicted optical flow features to produce new reference features $f^{\prime}_{ref}$. $f^{\prime}_{ref}$ are fed into a fusion module $\phi_2$, where inter-frame attention is calculated. This fusion sequentially predicts new facial features $f_{pre}$, with each frame's generation relying on all preceding reference information. Fusion $\phi_1$ consist of two linear layers and fusion $\phi_2$ is a 6-layer Transformer with a hidden layer dimension of 512. The use of an upper triangular mask is employed to prevent preceding frames from attending to future reference information. 
\begin{equation}
	f_{ref}= {\phi_1}(f_{face} \oplus f_{none}) ,\\ \enspace
	f_{pre}= {\phi_2}((f_{ref} \oplus f_{flow}) 
\end{equation}
\paragraph{Facial reconstruction}
The predicted facial features $f_{pre}$ are input into a decoder that learns to generate facial images. The network for facial reconstruction employs a U-net architecture, in which facial images are restored through upsampling. The predicted features are skip-connected with the reference features at various scales, aiming to preserve all the detail information from the reference video as much as possible.

\subsection{Optical Flow Synchronization Module}
Smooth facial movements and synchronized audio-frame changes are crucial metrics for evaluating video generation, particularly in talking face generation. It is essential that the generated sequence of facial frames adheres to the typical natural variations to enhance both realism and comprehensibility. Specifically, changes in lip movements should be consistent with the audio and textual semantics. 

As depicted at the left of \cref{fig:network}.(a) , We employ the reconstruction loss $\mathcal{L}_{rec}$ and the sync loss $\mathcal{L}_{sync}$ \cite{prajwal2020lip} to respectively supervise the generation quality of facial images as well as the alignment between audio and image:
\begin{equation}
	\mathcal{L}_{\text{rec}} = \frac{1}{N} \sum_{i=1}^N \| I_i - \hat{I}_i \|_1,
\end{equation}
\begin{equation}
\mathcal{L}_{\text{sync}} =  \frac{1}{N} \sum_{i=1}^N-\log \left( \frac{v \cdot s}{\max(\|v\|_2 \cdot \|s\|_2, \varepsilon)} \right),
\end{equation}
$I_i$ is groundtruth image and $ \hat{I}_i$ is generated image. The generator is trained to minimize L1 reconstruction loss between the generated frames $\hat{I}$ and ground-truth frames $ {I}$. $\mathcal{L}_{\text{sync}}$ is the loss of the expert lip-sync discriminator \cite{prajwal2020lip}  trained on the LRS2 dataset \cite{afouras2018deep}, where $v$ and $s$ are extracted by the speech encoder and image encoder in SyncNet \cite{chung2017out}.

As previously mentioned, the changes in pixel points within an image sequence during an extremely short time window should adhere to a smoothly regulated optical flow field. This is the foundation for natural facial movement and accurate lip-reading recognition. However, the "expert sync-loss" $\mathcal{L}_{\text{sync}}$ represents an average outcome across all image-audio pairs within a window and lacks continuity constraints for inter-frame variations. To address alignment requirements, generators often force images to undergo irrelevant action changes, resulting in visual illusions. Perceptually, this is evidenced by rapid, audio-independent movements of the synthesized lips, which is particularly apparent with slow-speaking audio. To mitigate this issue, we introduce an optical flow synchronization loss $\mathcal{L}_{\text{ofs}}$, hoping to eliminate these meaningless movements by supervising the consistency of optical flow changes and adding the continuity constraints between frames. Thanks to the development of optical flow computation models \cite{ilg2017flownet,jiang2021learning,teed2020raft} , we are now able to directly and rapidly calculate the optical flow information of consecutive frames. As shown at the right of \cref{fig:network}.(c) , we employ Global Motion Aggregation module \cite{jiang2021learning} to model the appearance self-similarities in reference facial frame and aggregate motion features, then calculate the optical flow between consecutive frames within a sequence window:
\begin{equation}
	o_{m,n} = \theta (GMA( \sigma(i_m),\tau(i_m, i_n)) \oplus \sigma(i_m) \oplus \tau(i_m, i_n))
\end{equation}
$o_{m,n}$ represents the optical flow between the image $i_m$ and the image $i_n$, $\sigma$ is a context (appearance) encoder, $\tau$ is a motion encoder. and $\theta$ is a GRU. $\oplus$ represents the concatenation of features. GMA and the specific details of optical flow estimation is introduced from \cite{jiang2021learning}.
In talking face videos, there is a noticeable disparity in the intensity of changes among features such as facial features, hair, and background elements. The lip area, in particular, exhibits changes in intensity that far surpass those of other regions. As such, calculating the average optical flow loss for an entire image is unreasonable. To better balance local and global optical flow changes, we generated separate optical flow for the lips and the whole image. This approach allows for a better restoration of lip variations and enhances the visual effect of the synthesis. During training, we minimize L2 optical flow synchronization loss $\mathcal{L}_{\text{ofs}}$ between the generated frames $\hat{I}$ and ground-truth frames $ {I}$:
\begin{equation*}
		\mathcal{L}_{\text{ofs-f}} = \frac{1}{N} \sum_{i=1}^N \| O_i - \hat{O}_i \|_2, \enspace \enspace \\
		\mathcal{L}_{\text{ofs-m}} = \frac{1}{N} \sum_{i=1}^N \| O_i^{m} - \hat{O}_i^{m} \|_2
\end{equation*}
\begin{equation}
		\mathcal{L}_{\text{ofs}} = (1-\alpha)\mathcal{L}_{\text{ofs-f}} +\alpha \mathcal{L}_{\text{ofs-m}} 
\end{equation}

$O_m$ is the ground-truth lip optical flow and $\hat{O}_i^{m}$ is the generated lip optical flow. The $\alpha$ is a balance hyperparameter that is typically set to 0.1.
Given the loss weights $\lambda$s, the total loss $L_{total}$ is represented as:
\begin{equation}
	\mathcal{L}_{\text{total}} = \lambda_1 \mathcal{L}_{\text{rec}} + \lambda_2 \mathcal{L}_{\text{sync}} +\lambda_3 \mathcal{L}_{\text{ofs}}
\end{equation}

\begin{table}[ht]
    \caption{Quantitative comparisons of Different Methods on LRS2 and HDTF Datasets}
    \label{Quantitative-table}
	\centering
	\resizebox{\linewidth}{!}{
		\begin{tabular}{@{}lcccccccccccccc@{}}
			\toprule
			& \multicolumn{6}{c}{LRS2 [61]} & \multicolumn{6}{c}{HDTF [75]} \\
			\cmidrule(lr){2-7} \cmidrule(lr){8-13}
			\textbf{Method} & \textbf{PSNR$\uparrow$} & \textbf{SSIM$\uparrow$} & \textbf{M/F-LMD$\downarrow$} & \textbf{FID$\downarrow$} & \textbf{LES-C$\uparrow$} & \textbf{LPIPS$\downarrow$} & \textbf{PSNR$\uparrow$} & \textbf{SSIM$\uparrow$} & \textbf{M/F-LMD$\downarrow$} & \textbf{FID$\downarrow$} & \textbf{LES-C$\uparrow$} &\textbf{LPIPS$\downarrow$} \\
			\midrule
			MakeItTalk \cite{zhou2020makelttalk} & 29.339 & 0.522 & 5.470/20.750 & 8.065 & 5.176 & 0.241 & 29.356 & 0.608 & 16.494/56.769 & 31.187 & 0.620 & 0.347 \\
			Wav2Lip \cite{prajwal2020lip} & 31.151 & 0.853 & 1.247/\underline{1.196} & 4.738 & 7.519 & \underline{0.099} & 29.756 & \underline{0.677} & \underline{3.369/2.708} & 23.514 & \textbf{4.632} & 0.275 \\
			Audio2Head \cite{wang2021audio2head} & 28.473 & 0.380 & 7.011/21.603 & 18.878 & 2.285 & 0.351 & 28.614 & 0.548 & 12.4237/42.8804 & 33.882 & 1.959 & 0.421 \\
			PC-AVS \cite{zhou2021pose} & 29.304 & 0.566 & 4.683/16.089 & 8.364 & 7.487 & 0.235 & 27.899 & 0.468 & 12.692/44.089 & 187.161 & 3.086 & 0.688 \\
			VideoReTalking \cite{cheng2022videoretalking} & 30.772 & 0.798 & 1.544/1.760 & 4.124 & 7.481 & 0.114 & \underline{29.866} & 0.673 & 3.993/3.637 & \underline{16.401} & 3.988 & 0.258 \\
			Sadtalker \cite{zhang2023sadtalker} & 29.316 & 0.526 & 5.682/19.651 & \underline{4.020} & 6.251 & 0.230 & 29.255 & 0.599 & 16.026/57.615 & 18.283 & \underline{4.488} & \underline{0.335}\\
			TalkLip \cite{wang2023seeing} & \underline{30.6039} & \underline{0.791} & \underline{1.2332}/1.5283 & 7.172 & \textbf{8.872} & \underline{0.138} & 29.619 & 0.659 & 7.243/21.962 & 25.736 & 3.367 & 0.302 \\
			Dreamtalker \cite{ma2023dreamtalk}] & 28.7058 & 0.431 & 8.2074/28.7102 & 26.796 & 2.479 & 0.318 & 28.727 & 0.557 & 10.856/39.030 & 31.363 & 2.325 & 0.406 \\
			Synctalk*  \cite{peng2023synctalk} & - & - & - & - &- & - & 28.031 & 0.487 & 5.795/9.765 & 51.783 & 3.2939 & 0.522\\
			Ours  & \textbf{32.832} & \textbf{0.888} & \textbf{0.890/0.960} & \textbf{2.635} & \underline{8.162} & \textbf{0.078} & \textbf{30.498} & \textbf{0.685} & \textbf{2.414/2.120} & \textbf{14.246} & 3.975 & \textbf{0.240}\\
			\midrule
			Ground Truth  & 100 & 1 & 0/0 & 0 & 8.248 & 0 & 100 & 1 & 0/0 & 0 & 3.931 & 0 \\
			\bottomrule
		\end{tabular}	}
\end{table}
\begin{figure}[htbp]
	\centering
	\includegraphics[width=\linewidth]{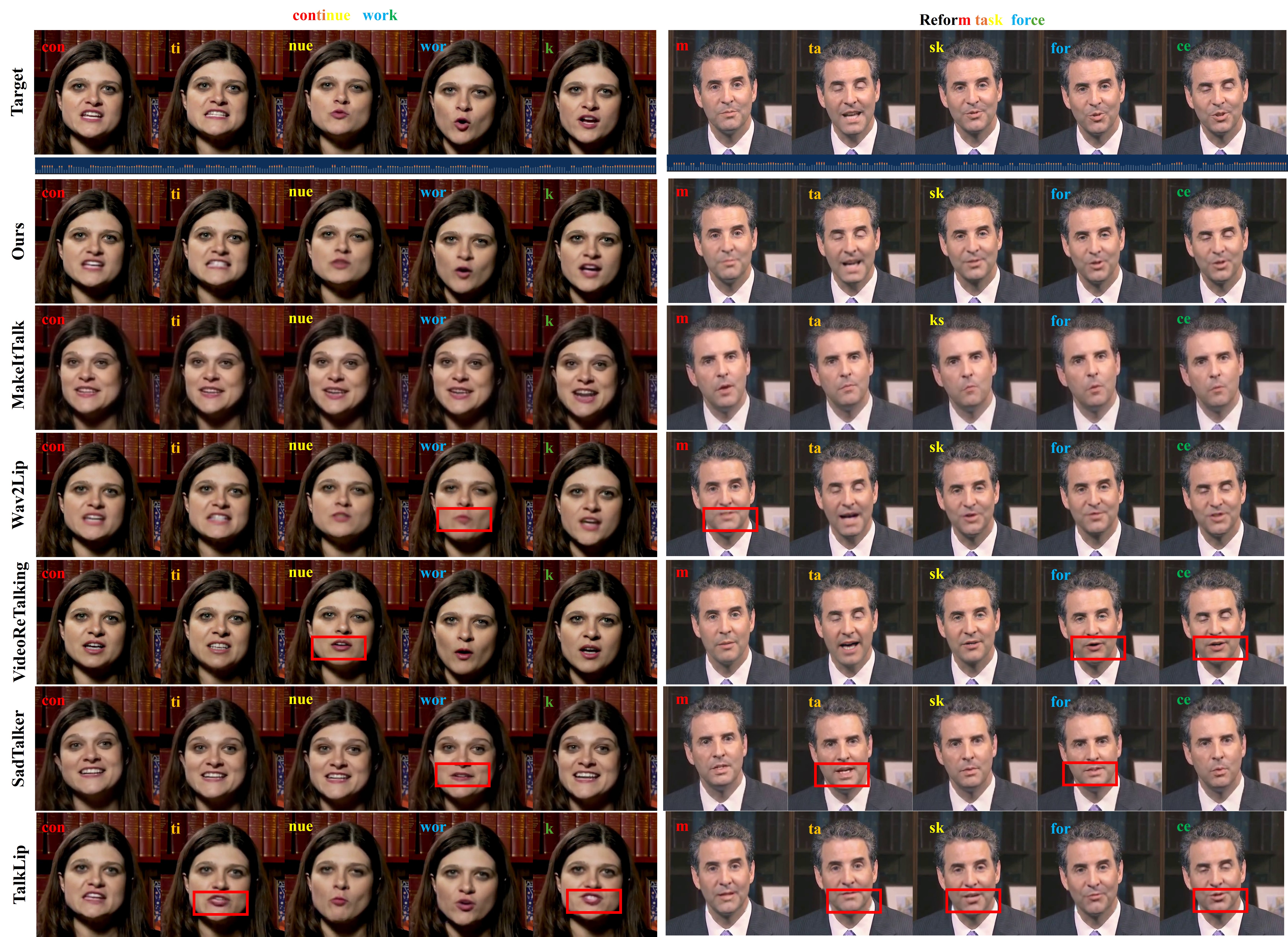}
	\caption{We compare our method with several state-of-the-art methods for audio-driven talking face generation. Different colors represent different syllables, corresponding to each image. }
	\label{fig:Qualitative}
\end{figure}
\section{Experiments}
\label{Results and Applications}
\subsection{Datasets and Experimental Settings}
\label{Datasets and Experimental Settings}
\paragraph{Datasets and Implementation Details} 
Our proposed method is trained and evaluated on the the datasets LRS2 \cite{afouras2018deep} and HDTF \cite{zhou2019talking}. The LRS2 dataset, which is utilized for training, contains over 45839 videos of 1251 different speakers. The test set includes 1243 videos. The HDTF dataset contains high-definition videos (15.8 hours) of 362 different individuals from YouTube. We randomly select 500 5-second clips as the test set for evaluation \cite{tan2024flowvqtalker}. Following previous methods \cite{siarohin2019first}, we crop the original videos and resize all data as a resolution of 256x256. Our method is trained using the Adam optimizer with a learning rate of $1e^{-4}$ and 256 batch size on 4 NVIDIA GeForce GTX 3090 GPUs or 1 NVIDIA Tesla A800 GPU . The default hyperparameters
are $\lambda_1 = 1$, $\lambda_2 = 0.1$, $\lambda_2 = 1$. Other training parameters are set following the Wav2Lip model.
\paragraph{Comparison Setting} 
We compare our method against state-of-the-art (SOTA) methods including neutral talking face generation methods: MakeItTalk \cite{zhou2020makelttalk}, Wav2Lip \cite{prajwal2020lip}, Audio2Head \cite{wang2021audio2head}, PC-AVS \cite{zhou2021pose}, VideoReTalking \cite{cheng2022videoretalking}, Sadtalker \cite{zhang2023sadtalker}, Talklip \cite{wang2023seeing}, Dreamtalker \cite{ma2023dreamtalk}, Synctalk \cite{peng2023synctalk}. Comparisons are evaluated on the following
widely used metrics: (1) generated video quality using PSNR, SSIM \cite{wang2004image}, LPIPS \cite{zhang2018unreasonable} and FID \cite{Seitzer2020FID}. (2) audio-visual synchronization using Landmarks Distances on the Mouth (M-LMD) \cite{chen2019hierarchical}, Landmarks Distances on the Face (F-LMD) ,  and the confidence score of SyncNet(LSE-C) \cite{prajwal2020lip}. (3) Lip-reading intelligibility: Visual Text Consistency Score (VTCS). VTCS employs two models: an audio transcription model and a lip-reading translation model \cite{ma2022visual,shi2022avhubert,shi2022avsr}, which separately identify the textual content depicted in the video. The consistency between the outputs of these models is measured using well-established linguistic metrics such as BLEU \cite{papineni2002bleu}, Word Error Rate(WER) \cite{ma2022visual}, Character Error Rate(CER), and ROUGE-L \cite{lin2004rouge}, which are commonly utilized in natural language translation to assess accuracy and coherence. Additionally, to account for the limitations of recognition models, we implement VTCS-B and VTCS-W, which specifically measure the discrepancy in visual-semantic alignment between synthesized videos and actual samples, providing a more comprehensive evaluation of the synthetic video's realism:
\begin{equation*}
	VTCS-B= \frac{1}{N}\sum_{1}^{n}\|BLEU(T_{gen}, T_{speech})-BLEU(T_{gt}, T_{speech})\|_{2} \enspace \enspace \\
\end{equation*}
\begin{equation}
	VTCS-W= \frac{1}{N}\sum_{1}^{n}\|WER(T_{gen}, T_{speech})-WER(T_{gt}, T_{speech})\|_{2} \enspace \enspace \\
\end{equation}
$T_{gen}$, $T_{gt}$, $T_{speech}$ represent the text from generated videos, the text from ground-truth videos, the text from speeches. Partial results are moved to Appendix due to limited space.

\begin{table}[ht]
	\centering
	\caption{Lip-reading intelligibility comparisons of Different Methods on LRS2 Datasets}
	\label{Lip-table}
	\resizebox{0.8\linewidth}{!}{
		\begin{tabular}{@{}lcccccccc@{}}
			\toprule
			\textbf{Method} & \textbf{BLEU$\uparrow$} & \textbf{WER$\downarrow$} & \textbf{CER$\downarrow$} & \textbf{ROUGE-L$\uparrow$} & \textbf{VTCS-B$\downarrow$} & \textbf{VTCS-W$\downarrow$} \\
			\midrule
			MakeItTalk \cite{zhou2020makelttalk} & 0.0349 & 1.1345 & 0.8232 & 0.0689 & 0.0242 & 0.0245 \\
			Wav2Lip \cite{prajwal2020lip} & 0.1798 & 0.9915 & 0.6674 & 0.1526 &	0.0207 & 0.0209 \\
			Audio2Head \cite{wang2021audio2head} & 0.0448 & 1.1451 & 0.8074 & 0.0512 & 0.0238 & 0.0246 \\
			PC-AVS \cite{zhou2021pose} & 0.1185 & 1.0320 & 0.7120 & 0.1128 & 0.0256 & 0.0250 \\
			VideoReTalking \cite{cheng2022videoretalking} & 0.2448 & 0.9196 & 0.6078 & 0.2044 & 0.0198 & 0.0198  \\
			Sadtalker \cite{zhang2023sadtalker} & 0.0746 & 1.1063 & 0.7726 & 0.0837 & 0.0231 & 0.0237\\
			Dreamtalker \cite{ma2023dreamtalk}] & 0.0904 & 1.0744 &0.7463 & 0.0945 & 0.0228 & 0.0229 \\
			Ours  & \textbf{0.3273} & \textbf{0.8275} & \textbf{0.5326} & \textbf{0.2718} & \textbf{0.0175} & \textbf{0.0174} \\
			\midrule
			Ground Truth  & 0.8401 & 0.3508 & 0.1383 & 0.6089 & 0 & 0 \\
			Talklip* \cite{wang2023seeing} & 0.7601 & 0.4253 & 0.1927 & 0.5634 & 0.009 & 0.0088 \\
			\bottomrule
	\end{tabular}	}
\end{table}
\subsection{Experimental Results}
\paragraph{Quantitative Results} 
The quantitative results are presented in Table.\ref{Quantitative-table}. where our OpFlowTalker achieve the best performance across most metrics, except LES-C. Wav2Lip and Sadtalker pretrains the SyncNet discriminator directly, which might lead them to prioritize achieving a higher confidence score of SyncNet over optimizing visual performance, such as the unnatural lip movements. It is evident in their worse Lip-reading intelligibility in Table.\ref{Lip-table} and inferior M-LMD score to our method. Our method achieves a similar score to the real video, which demon strates our advantage. Talklip adopts a lip-reading experter pre-trained on the LRS2 dataset, which might enhance the lip-reading intelligibility on this, but the lower video quality (FID: 7.172) and poor performance on other datasets(LES-C: 3.367, M/F-LMD: 7.243/21.962 on HDTF) indicate a lack of generalizability. Synctalk requires high-resolution reference images for synthesis, making it unsuitable for testing on the LRS2 dataset. The Lip-reading intelligibility results are presented in Table.\ref{Lip-table}. Compared to all methods except TalkLip, our method has shown significant improvements across all metrics, underlining its effectiveness in enhancing the lip-readability of synthesized videos. During its development, TalkLip utilizes a lip-reading model \cite{shi2022avhubert} fine-tuned on the LRS2 dataset for supervision, which enables it to achieve far superior lip-readability compared to all other methods without lip-reading supervision.
\paragraph{Qualitative results}
To qualitatively evaluate different methods, we provide the visual demonstrations of two generated talking face videos in \ref{fig:Qualitative}.Specifically, the first row provides the ground truth video, followed by image samples synthesized by different methods. From Figure.\ref{fig:Qualitative}, it can be seen that our method provides image frames most similar to the real video, with smooth transitions in mouth shapes. In Figure.\ref{fig:Qualitative}, our approach also performs best in terms of video quality and the naturalness of mouth movements. For more qualitative results, please refer to the appendix.\ref{More Qualitative results}.

\subsection{Ablation study}	
\label{Ablation study}

We conduct ablation experiments on different modules of the proposed method, as shown in Table.\ref{ablation-table}. We test the impact of different optical flow estimation methods: FlowNet2 \cite{ilg2017flownet}, RAFT \cite{teed2020raft}, and GMA \cite{jiang2021learning} on the quality of the generated videos. The time taken to estimate optical flow for a single image on a single RTX 3090 GPU is as follows: 0.12s, 0.05s, and 0.06s. In the calculation of optical flow estimation loss for facial images, aggregating global motion information slightly increases the FID score. However, it significantly improves inter-frame continuity and semantic consistency. While a higher FID generally suggests a deviation from the reference dataset's distribution, the improved continuity and consistency achieved through capturing global motion are more vital for generating realistic and seamless videos. As shown in Table.\ref{ablation-table}.b, increasing the interval between frames for optical flow estimation may slightly improve the FID score, but negatively affect the quality and semantic consistency of the synthesized videos. A larger frame interval means that the perception of motion updates is less frequent, which may cause the system to overlook finer and quicker changes. This potential gap in capturing detailed movements can result in a less accurate and coherent portrayal of motion in the synthesized videos, making the overall video appear less lifelike and potentially disjointed. Table.\ref{ablation-table}.c shows the results of using full-face optical flow, lip optical flow, and both simultaneously. Compared to using just one type of optical flow, employing both types may slightly increase the LMD score, but overall performance is improved. This suggests that integrating both full-face and lip optical flows enhances the accuracy and naturalness of facial movements in synthesized videos, providing a more comprehensive capture of facial dynamics.The results comparing our proposed components with the baseline are displayed in Table.\ref{ablation-table}.d, illustrating the effectiveness of our method. More ablation studies are presented in the appendix.\ref{More Ablation Study}.

\begin{table}[ht]
	\centering
	\caption{Ablation study results of different components on LRS2 datasets}
	\label{ablation-table}
	\resizebox{\linewidth}{!}{
		\begin{tabular}{@{}lccccccccc@{}}
			\toprule
			\textbf{Comparison} & \textbf{Method} & \textbf{PSNR$\uparrow$} & \textbf{SSIM$\uparrow$} & \textbf{FID$\downarrow$} & \textbf{LSE-C$\uparrow$} & \textbf{M/F-LMD$\downarrow$} & \textbf{ROUGE-L$\uparrow$} & \textbf{VTCS-B$\downarrow$} & \textbf{VTCS-W$\downarrow$} \\
			\midrule
			Baseline & Wav2Lip \cite{prajwal2020lip} & 31.150 & 0.853 & 4.739 & 7.519 & 1.247/1.196 & 0.1526 & 0.0207 & 0.0209 \\
			\midrule
			\multirow{3}{*}{a: Optical flow estimation} &
			FlowNet2 \cite{ilg2017flownet} 	& 32.384 & 0.881 & \textbf{2.529}& 7.978 & 0.892/0.963 & 0.2590 & 0.0179 & 0.0176 \\
			& RAFT \cite{teed2020raft} 		& 32.488 & 0.886 & 2.570         & 7.923 & 0.910/0.966 & 0.2643 & 0.0177 & 0.0175 \\
			& GMA \cite{jiang2021learning}	& \textbf{32.832}& \textbf{0.888} & 2.635 & \textbf{8.162} & \textbf{0.890/0.960} & \textbf{0.2718} & \textbf{0.0175} & \textbf{0.0174} \\
			\midrule
			\multirow{3}{*}{b: Inter-frame interval (GMA) } &
			1  & \textbf{32.832}& \textbf{0.888} & 2.635 & \textbf{8.162} & \textbf{0.890/0.960} & \textbf{0.2718} & \textbf{0.0175} & \textbf{0.0174} \\
			& 2  & 32.389 & 0.885 & \textbf{2.614} & 8.002 & 0.950/0.995 & 0.2571 & 0.0179 & 0.0181  \\
			& 5  & 32.796 & 0.886 & 2.662 & 8.061 & 0.920/0.977 & 0.2555 & 0,0181 & 0.0180  \\
			\midrule
			\multirow{3}{*}{c: Optical flow area} &
			Face & 32.376 & 0.882 & \textbf{2.721} & 8.124 & \textbf{0.956/1.007} & 0.2495 & 0.0181 & 0.0180 \\
			& Mouth & 32.388 & 0.880 & 2.927  & 8.276 & 0.971/1.010 & 0.2492 & 0.0182 & 0.0183 \\
			& Face+Mouth & \textbf{32.638}& \textbf{0.883} & 2.765 & \textbf{8.320} & 0.959/1.014 & \textbf{0.2561} & \textbf{0.0178} & \textbf{0.0177} \\
			\midrule
			\multirow{4}{*}{d: modules} &
			Audio Augmentation & 32.547 & 0.883 & 2.911 & 8.138 & 0.971/1.005 &  0.2251 & 0.0188 & 0.0188 \\
			& Optical Flow predictor & 32.588 & 0.881 & 2.694 & \textbf{8.237} & 0.985/1.015 & 0.2426 & 0.0184 & 0.0183  \\
			& Sequential Fusion  & 32.583 & 0.884 & 2.669 & 7.800 & 0.946/0.996 & 0.2325 & 0.0188 & 0.0189  \\
			& OFSM  & 32.371 & 0.876 & 2.849 & 8.200 & 1.010/1.043 & 0.2508 & 0,0181 & 0.0182  \\
			& Full model & \textbf{32.832}& \textbf{0.888} & \textbf{2.635} & 8.162 & \textbf{0.890/0.960} & \textbf{0.2718} & \textbf{0.0175} & \textbf{0.0174} \\
			\midrule
			Ground Truth  & - & 100 & 1 & 0 & 8.248 & 0/0 & 0.6089  & 0 & 0 \\
			\bottomrule
	\end{tabular}	}
\end{table}
\section{Conclusion}
We have systematically introduced optical flow into the task of talking face generation for the first time, proposing OpFlowTalker, a novel method designed to generate videos by predicting optical flow changes from audio instead of directly predicting portrait images. OpFlowTalker samples predicted optical flow changes from an enhanced audio distribution, uses sequence fusion to perceive contextual information, and utilizes inter-frame optical flow supervision of both the full face and lips to guide the learning of the generator. By learning about potential changes rather than fixed facial mappings, we have improved the inter-frame continuity, semantic consistency, and enhanced the generality of the model. Our experiments demonstrate the superiority of our method in synthesizing audio-visually aligned, lip-readable, high-quality portrait videos, offering insightful implications for subsequent related work.
We discuss the limitations of the work in the section.\ref{Limitation}.


%
%



\bibliographystyle{splncs04}
\bibliography{neurips_2024}
\newpage

\appendix

\section{Appendix / supplemental material}

Optionally include supplemental material (complete proofs, additional experiments and plots) in appendix.
All such materials \textbf{SHOULD be included in the main submission.}
\subsection{Quantitative Results}
\label{More Quantitative Results}
\begin{table}[ht]
	\centering
	\caption{Lip-reading intelligibility comparisons of Different Methods on HDTF Datasets}
	\label{Lip-HDTF-table}
	\resizebox{\linewidth}{!}{
		\begin{tabular}{@{}lcccccccc@{}}
			\toprule
			\textbf{Method} & \textbf{BLEU$\uparrow$} & \textbf{WER$\downarrow$} & \textbf{CER$\downarrow$} & \textbf{ROUGE-L$\uparrow$} & \textbf{VTCS-B$\downarrow$} & \textbf{VTCS-W$\downarrow$} \\
			\midrule
			Ours & \textbf{0.3389} & \textbf{0.8542} & \textbf{0.5465} & \textbf{0.3530} & \textbf{0.0269} & \textbf{0.0105} \\
			Talklip* \cite{wang2023seeing} & 0.2001 &	1.033 & 0.6881 & 0.2104	& 0.0242 & 0.0329 \\
			\midrule
			Ground Truth  & 0.7011 & 0.3849 & 0.2161 & 0.7242 & 0 & 0 \\
			\bottomrule
	\end{tabular}	}
\end{table}
We evaluated the lip-readability of synthesized videos by different models on the HDTF dataset, as shown in Table.\ref{Lip-HDTF-table}. During the training of TalkLip, a lip-reading recognition model trained on the LRS2 dataset was used as one of the loss components for the discriminator. This resulted in highly effective lip-reading performance on the LRS2 dataset. However, results on the HDTF dataset indicate that this lip-reading capability lacks generalizability. In contrast, our method continued to achieve the best results, demonstrating its robustness and superior performance across different datasets. This highlights our approach's adaptability and effectiveness in producing high-quality, lip-readable talking face videos in various contexts.

\subsection{Qualitative results}
\label{More Qualitative results}
\begin{figure}
	\centering 
	\includegraphics[width=\linewidth]{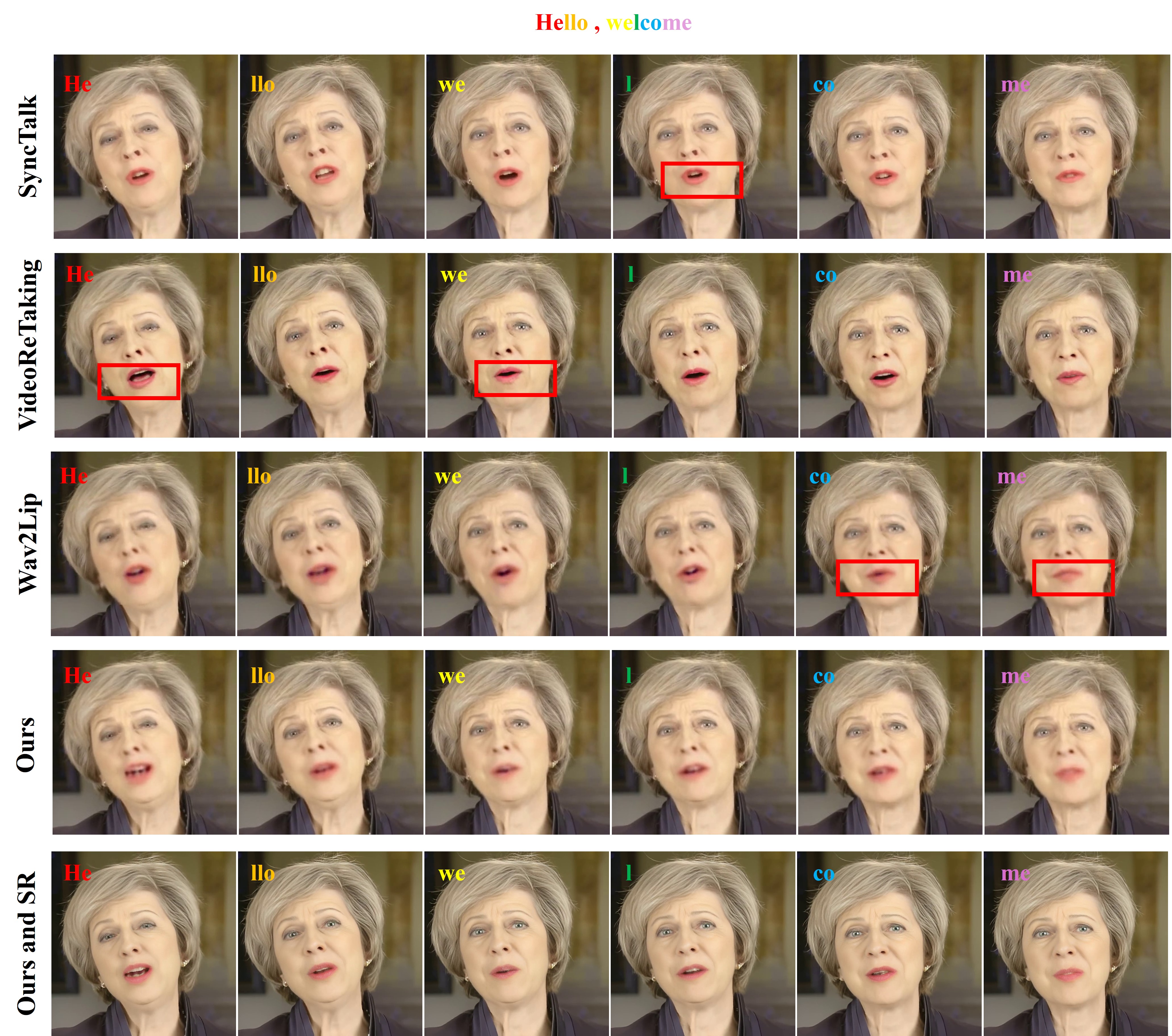}
	\caption{A comparison of our method with Synctalk, DreamReTalking, and Wav2Lip in generating high-definition videos (256x256) on the HDTF dataset. SR stands for Super-Resolution. }
	\label{fig:Qualitative2}
\end{figure}

\begin{figure}
	\centering
	\includegraphics[width=\linewidth]{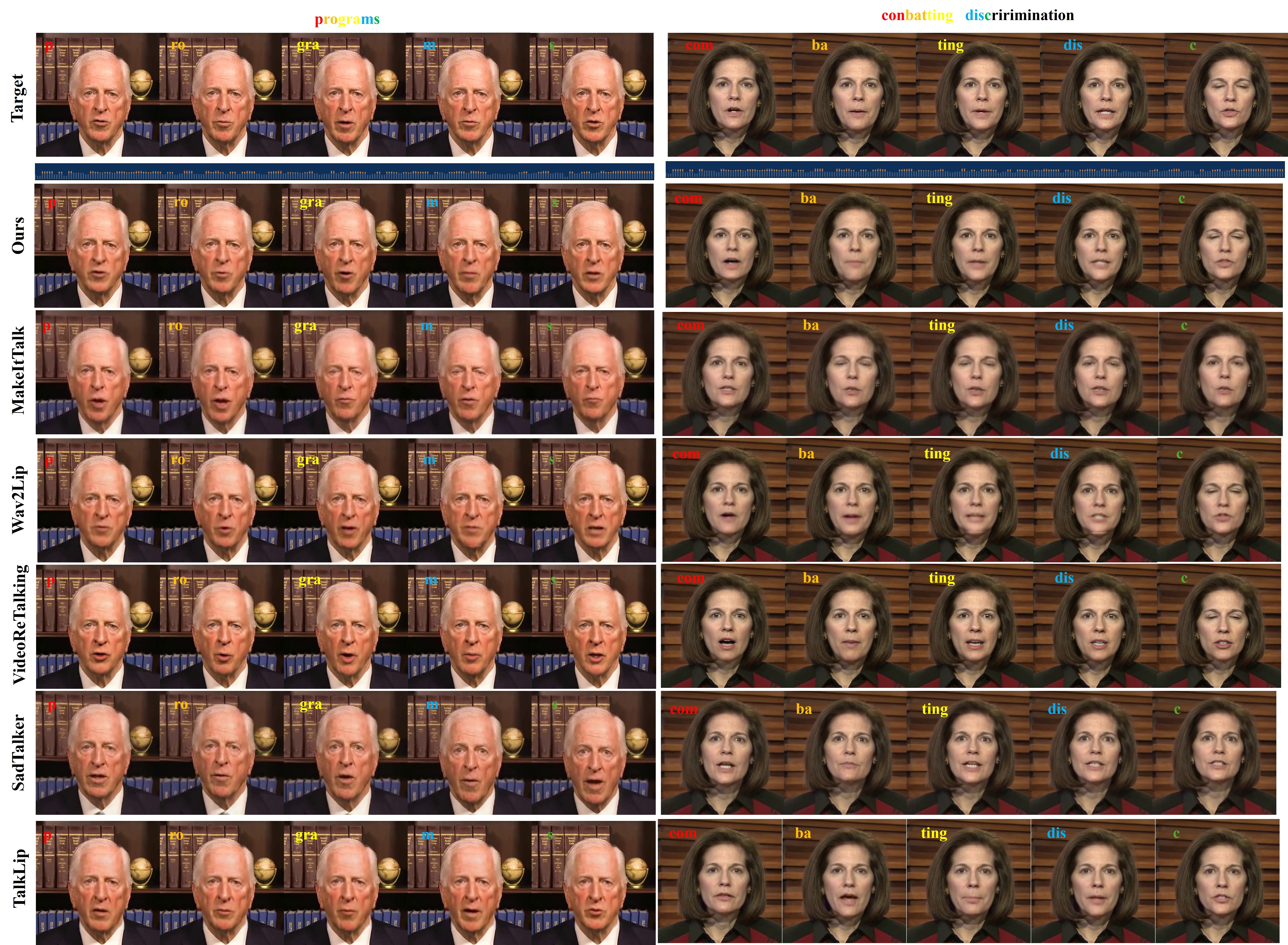}
	\caption{More comparisons of our method with several state-of-the-art methods for audio-driven talking face generation. Different colors represent different syllables, corresponding to each image. }
	\label{fig:Qualitative3}
\end{figure}

\begin{figure}
	\centering
	\includegraphics[width=\linewidth]{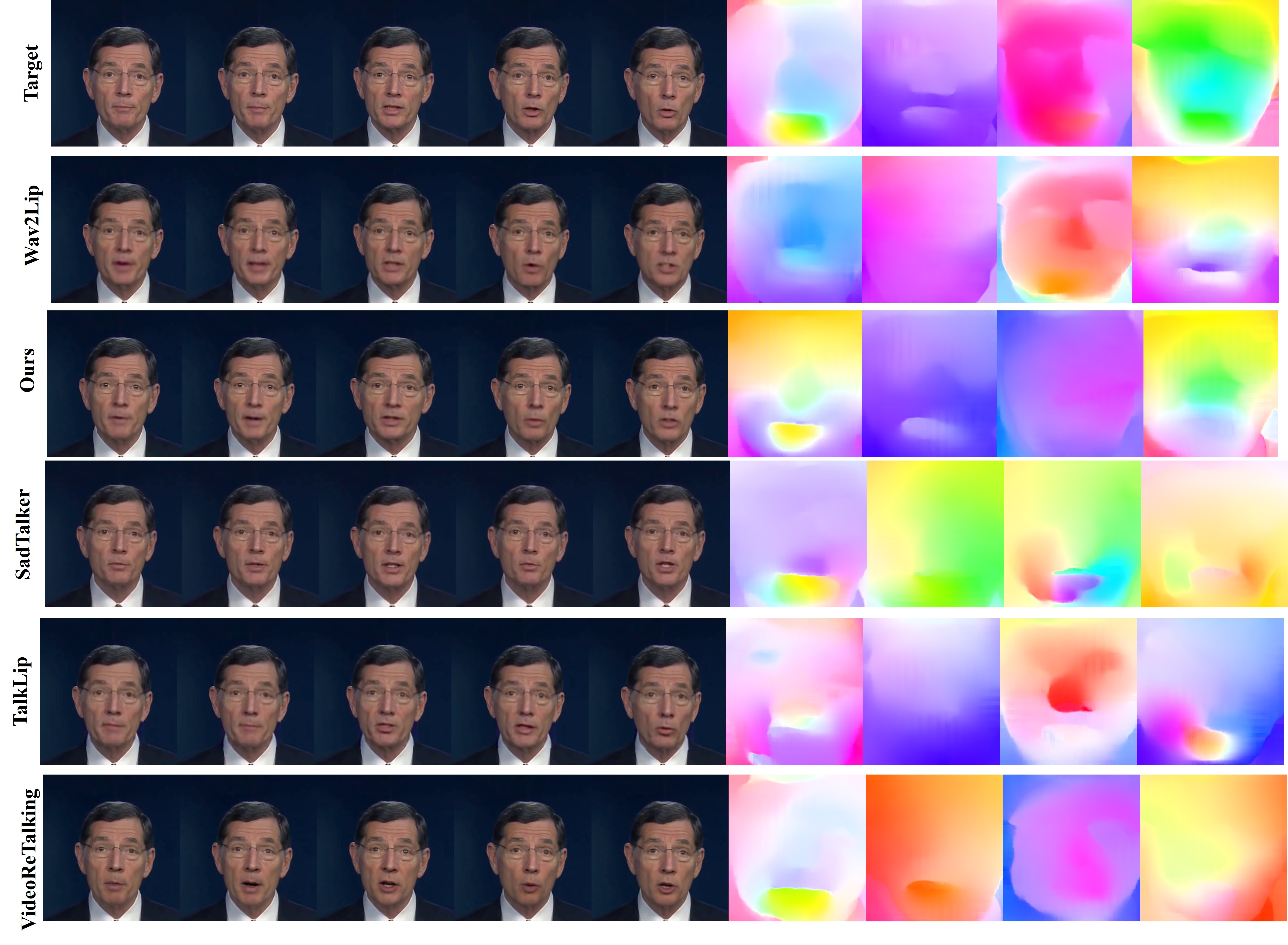}
	\caption{A comparison of the optical flow fields generated by our method, other methods, and those from real videos. }
	\label{fig:Qualitative4}
\end{figure}

Figure.\ref{fig:Qualitative2} presents a comparison of our method with Synctalk, DreamReTalking, and Wav2Lip in generating high-definition videos (256x256) on the HDTF dataset. Synctalk employs a customized training approach for individual characters, which, although effective, limits its generalizability since each new character image necessitates training a new model. DreamReTalking, on the other hand, boasts superior video clarity due to its training on a substantial high-definition dataset, but it underperforms in audio-lip synchronization. In contrast, our method, despite being trained only on the lower resolution LRS2 dataset (96x96), performs comparably to models trained on specific or extensive HD datasets, particularly excelling in audio-lip synchronization. This demonstrates the generalizability and effectiveness of our approach. Notably, our method does not require individualized training and can adapt to video generation needs across varying resolutions. Figure.\ref{fig:Qualitative3} provides additional visual effect comparisons. Figure.\ref{fig:Qualitative4} presents a comparison of the optical flow fields generated by our method, other methods, and those from real videos. The results indicate that the optical flow changes produced by our method are closer to those of real videos, demonstrating its high precision and effectiveness in simulating authentic facial dynamics. This result underscores our method's ability to accurately capture and reproduce the subtle movements involved in natural human expressions, thereby enhancing the realism and believability of the generated video content. Our method is particularly significant for applications in virtual reality, film production, and other areas where lifelike facial animations are crucial. The supplementary materials which include video samples, further highlight the visual effectiveness and superiority of our approach.

\subsection{Ablation Study} 
\label{More Ablation Study}
Table.\ref{ablation-HDTF-table} presents the results of ablation experiments on the HDTF dataset. GMA still offers the best overall quality, while FlowNet2 performs better in terms of lip alignment. This is because when aggregating global motion information for rapid optical flow estimation, there is a certain degree of loss in the precision of local details. The choice between the two should consider the clarity of the dataset and practical requirements comprehensively. The selection of frame intervals for optical flow estimation has its pros and cons. Because the speech rate in HDTF is relatively steady and facial changes are more subtle, increasing the interval frames for optical flow calculation can maintain inter-frame continuity while reducing the quality degradation caused by forced alignment. The computation of optical flow regions and module ablation results are consistent with those on the LRS2 dataset.

\begin{table}[ht]
	\centering
	\caption{Ablation study results of different components on HDTF datasets}
	\label{ablation-HDTF-table}
	\resizebox{\linewidth}{!}{
	\begin{tabular}{@{}lccccccccc@{}}
		\toprule
		\textbf{Comparison} & \textbf{Method} & \textbf{PSNR$\uparrow$} & \textbf{SSIM$\uparrow$} & \textbf{FID$\downarrow$} & \textbf{LSE-C$\uparrow$} & \textbf{M/F-LMD$\downarrow$} & \textbf{LPIPS$\downarrow$} \\
		\midrule
		Baseline & Wav2Lip \cite{prajwal2020lip} & 29.756 & 0.677 & 23.515. & 4.632 & 3.369/2.70 & 0.275\\
		\midrule
		\multirow{3}{*}{a: Optical flow estimation} &
		FlowNet2 \cite{ilg2017flownet} 	& 30.406 & 0.685 & \textbf{13.671}& 3.924 & \textbf{2.345/2.069} & \textbf{0.239} \\
		& RAFT \cite{teed2020raft} 		& 30.438 & 0.685 & 13.882         & 3.967 & 2.417/2.092 & 0.240  \\
		& GMA \cite{jiang2021learning}	& \textbf{30.498}& \textbf{0.685} & 14.246 & \textbf{3.975} & 2.414/2.120 & 0.241 \\
		\midrule
		\multirow{3}{*}{b: Inter-frame interval (GMA) } &
		1  & 30.498& 0.685 & 14.246 & 3.975 & \textbf{2.414/2.120} & 0.241 \\
		& 2  & 30.366 			& 0.684 & \textbf{14.020} & \textbf{4.108} & 2.594/2.230 & 0.243 \\
		& 5  & \textbf{30.515} & \textbf{0.686} & 14.163 & 3.982 & 2.489/2.176 & \textbf{0.239}   \\
		\midrule
		\multirow{3}{*}{c: Optical flow area} &
		Face & 30.306 & 0.684 & 15.248 & 4.144 & \textbf{2.619/2.296} & 0.243  \\
		& Mouth & 30.335 & 0.685 & 15.256  & 4.208 & 2.680/2.336 & 0.245  \\
		& Face+Mouth & \textbf{30.438}& \textbf{0.685} & \textbf{14.469} & \textbf{4.308} & 2.641/2.349 & \textbf{0.242}  \\
		\midrule
		\multirow{4}{*}{d: modules} &
		Optical Flow predictor & 30.455 & 0.684 & 15.082 & 4.144 & 2.727/2.313 & 0.242 \\
		& Sequential Fusion  & 30.393 & 0.685 & 14.664 & 4.000 & 2.528/2.249 & 0.243 \\
		& OFSM  & 30.336 & 0.682 & 15.664 & \textbf{4.281} & 2.956/2.517 & 0.245   \\
		& Full model & \textbf{30.515}& \textbf{0.686} & \textbf{14.163} & 3.982 & \textbf{2.489/2.176} & \textbf{0.239} \\
		\midrule
		Ground Truth  & - & 100 & 1 & 0 & 3.932 & 0/0 & 0  \\
		\bottomrule
\end{tabular}	}
\end{table}
Table.\ref{Mse-ablation-table} shows a comparison of optical flow loss calculation methods. MSE (Mean Squared Error) performs better in terms of video quality and lip-readability, while MAE (Mean Absolute Error) excels in FID (Fréchet Inception Distance), LPIPS (Learned Perceptual Image Patch Similarity), and audio-lip alignment. Because MSE is more robust and provides good results across various videos, it is generally preferred for ensuring overall quality and consistency. On the other hand, MAE is better at handling outliers and excels in learning feature distances and alignment. The choice of loss function should be tailored to the specific priorities of the application, whether focusing on visual quality, perceptual similarity, or alignment precision.

Table.\ref{max-ablation-table} illustrates the differences in training outcomes depending on whether the optical flow loss is calculated as the average discrepancy across all frames within a window or the maximum discrepancy. This comparison helps to determine the impact of using average versus maximum error measurements on the training process and the quality of the resulting synthesized video. Calculating the average discrepancy tends to promote consistency and smoothness across the entire sequence, whereas focusing on the maximum discrepancy can help in prioritizing the correction of the most significant errors in frame transitions, potentially leading to better alignment in the most challenging frames but possibly at the cost of overall smoothness
\subsection{Limitation}
\label{Limitation}	
While our method successfully generates realistic, natural, and semantically consistent videos from reference videos and audio, it does have limitations. A primary constraint is the low resolution of the training videos from the LRS2 dataset, which hampers the capture of optical flow and facial reconstruction capabilities, limiting performance on high-definition datasets. Enhancing performance on high-definition content may require training with higher-resolution image datasets. Additionally, our focus predominantly on lip movements and inter-frame changes other than the other facial expressions, such as emotions and gaze. Thus, the emotional content of the generated videos tends to be relatively static, detracting from the authenticity of the synthesized content. Addressing the inclusion of a broader range of facial expressions represents a significant direction for future research. 
\subsection{Broader Impact}
\label{Societal Impacts}
Our method demonstrates exceptional effectiveness in the task of talking face generation and holds significant potential across various application contexts, such as virtual reality\cite{peng2023selftalk}, film production\cite{kim2018deep}, and online education\cite{pataranutaporn2021ai}. Although there is a risk that the related models could be misused, the important social implications of this technology warrant controlled use and research support. Therefore, we plan to restrict access to the complete pre-trained models to prevent misuse of the methods and technologies. This approach balances the benefits of advancing this field with responsible management to mitigate potential risks.

\begin{table}[ht]
	\centering
	\caption{Different methods of the Optical FLow Loss calculation on LRS2 datasets}
	\label{Mse-ablation-table}
	\resizebox{\linewidth}{!}{
		\begin{tabular}{@{}lcccccccccccc@{}}
			\toprule
			\textbf{Method} & \textbf{PSNR$\uparrow$} & \textbf{SSIM$\uparrow$} & \textbf{FID$\downarrow$} & \textbf{LPIPS$\downarrow$} & \textbf{LSE-C$\uparrow$} & \textbf{LSE-D$\downarrow$} & \textbf{M/F-LMD$\downarrow$} & \textbf{ROUGE-L$\uparrow$} & \textbf{VTCS-B$\downarrow$} & \textbf{VTCS-W$\downarrow$} & \\
			\midrule
			MSE & \textbf{32.712} & \textbf{0.884} & 2.712 & 0.081 & 8.130 & 6.043 & \textbf{0.972/1.0146} & \textbf{0.2481} & \textbf{0.0182} & \textbf{0.0178} \\
			MAE & 32.588 & 0.881 & \textbf{2.694} & \textbf{0.080} & \textbf{8.237} & \textbf{6.020} & 0.9851/1.015 & 0.2426 & 0.0184 & 0.0183 \\
			Ground Truth  & 100 & 1 & 0 & 0 & 8.248 & 6.258 & 0/0 & 0.6089  & 0 & 0 \\
			\bottomrule
	\end{tabular}	}
\end{table}

\begin{table}[ht]
	\centering
	\caption{Different methods of the Optical FLow Loss in a windows on LRS2 datasets}
	\label{max-ablation-table}
	\resizebox{\linewidth}{!}{
		\begin{tabular}{@{}lcccccccccccc@{}}
			\toprule
			\textbf{Method} & \textbf{PSNR$\uparrow$} & \textbf{SSIM$\uparrow$} & \textbf{FID$\downarrow$} & \textbf{LPIPS$\uparrow$} & \textbf{LSE-C$\uparrow$} & \textbf{LSE-D$\downarrow$} & \textbf{M/F-LMD$\downarrow$} & \textbf{ROUGE-L$\uparrow$} & \textbf{VTCS-B$\downarrow$} & \textbf{VTCS-W$\downarrow$} & \\
			\midrule
			Mean & 32.376 & 0.882 & \textbf{2.721} & 0.080 & 8.124 & 6.036 & 0.956/1.007 & 0.2495 & 0.0181 & 0.0180 \\
			Max  & \textbf{32.523} & \textbf{0.883} & 2.824 & \textbf{0.075} & \textbf{8.300} & \textbf{5.905} & \textbf{0.943/0.996} & \textbf{0.2548} & \textbf{0.0180} & \textbf{0.0177} \\
			Ground Truth  & 100 & 1 & 0 & 0 & 8.248 & 6.258 & 0/0 & 0.6089  & 0 & 0 \\
			\bottomrule
	\end{tabular}	}
\end{table}
\end{document}